\documentclass[letterpaper]{article} 
\usepackage{aaai23}  
\usepackage{times}  
\usepackage{helvet}  
\usepackage{courier}  
\usepackage[hyphens]{url}  
\usepackage{graphicx} 
\usepackage{natbib}  
\usepackage{caption} 
\usepackage{algorithm}
\usepackage{algorithmic}

\usepackage{newfloat}
\usepackage{listings}
\DeclareCaptionStyle{ruled}{labelfont=normalfont,labelsep=colon,strut=off} 
\floatstyle{ruled}
\newfloat{listing}{tb}{lst}{}
\floatname{listing}{Listing}
\title{Bridging Generative Networks with the Common Model of Cognition}
\author{
    Robert L. West,
    Spencer Eckler,
    Brendan Conway-Smith,
    Nico Turcas,
    Eilene Tomkins-Flanagan,
    \\Mary Alexandria Kelly
}
\affiliations{
    Carleton University, Ottawa, ON, Canada\\
    Robert.West@carleton.ca,
    Spencer.Eckler@carleton.ca,
    Brendan.ConwaySmith@carleton.ca,
    Nico.Turcas@carleton.ca,
    Eilene.TomkinsFlanaga@carleton.ca,
    Mary.Kelly4@carleton.ca\\

}

\usepackage{bibentry}

\begin{document}

\maketitle

\begin{abstract}
This article presents a theoretical framework for adapting the Common Model of Cognition to large generative network models within the field of artificial intelligence. This can be accomplished by restructuring modules within the Common Model into shadow production systems that are peripheral to a central production system, which handles higher-level reasoning based on the shadow productions' output. Implementing this novel structure within the Common Model allows for a seamless connection between cognitive architectures and generative neural networks.
\end{abstract}

\section{Introduction}

Intelligent systems gain significant robustness by possessing both Good Old-Fashioned Artificial Intelligence (i.e., GOFAI or ``symbolic'') reasoning and connectionist statistical learning \citep[e.g.][]{hitzler2022neuro}; however, there is no consensus on how to integrate the two. One of the less explored methods involves integrating generative AI models and cognitive architectures into a single hybrid system. A leading candidate for modeling the architecture of human cognition is the Common Model of Cognition, formerly the Standard Model of the Mind \cite{laird2017standard}, however it currently lacks a method to make lower-level connectionist factors interpretable at the cognitive level. The Common Model of Cognition (CMC) provides an account of how human cognition operates computationally and has been validated by large-scale neuroscience data \cite{Stocco2021cmc}. In contrast, most generative neural networks are not constrained by correspondence to biology and instead take a pragmatic approach toward generating intelligent output. 

Cognitive modelling and artificial intelligence have distinct goals, namely to explain and predict the behaviour of humans and animals on the one hand, and to solve  problems and perform tasks without human guidance on the other. Nevertheless, cognitive models can benefit from the integration of current deep learning approaches, as many tasks solved by generative networks are tasks for which cognitive modelling lacks detailed process models, such as perception, imagination, and natural language processing. As such, large generative networks can be understood as candidate cognitive models for how humans accomplish these tasks. However, generative networks, such as large neural language models, have been shown to have significant limitations in formal reasoning abilities \cite{helwe2021shallow}, in contrast to traditional ``language and logic'' based approaches popular in cognitive models, designed to follow structured reasoning \cite{shaw1958logic} and problem solving processes \cite{west2007sgoms}. Integrating traditional cognitive modelling approaches with generative networks may therefore yield architectures that are better able to support modelling the full range of human behaviour, and may broaden the range of problems solvable by a unitary AI system.

To bridge deep learning with traditional cognitive modelling approaches, we propose significant adjustments to the Common Model of Cognition (CMC) to enhance CMC models with more advanced, large-scale cognitive abilities. We briefly discuss the Common Model of Cognition and the ACT-R cognitive architecture \cite{Anderson1998} and following this, how it can be reformulated. 

The central contribution of this paper is a proposed reformulation of how ACT-R conceives of modules. ACT-R holds that human cognition consists of modules that handle specific cognitive capacities, such as perception, motor, and language, as well as a central executive consisting of procedural memory and working memory. As there is no standard for how these modules should implement these cognitive capacities, ACT-R modules are, in practice, implemented in an \textit{ad hoc} manner. We propose a standardized approach for implementing modules in ACT-R and other CMC architectures, where modules consist of generative, pre-trained networks, a production system for managing networks, and a memory system that  interfaces between networks and production systems. 

\section{Components of the Common Model}

The Common Model represents a convergence across cognitive architectures as to the components necessary for human-like intelligence. The Common Model contains five modules: perception, working memory, motor, declarative memory, and procedural memory. Procedural memory is central to how the Common Model operates, as it responds to and instructs other modules based on the contents of working memory. A straightforward approach to constructing procedural memory is by way of a production system. A production system is a long established method of implementing the basic functions described by the Common Model, and hence we will refer to production systems with the caveat that more complex systems can also be used. This method is adopted in ACT-R, but differs from other Common Model Architectures like SOAR \cite{laird2012} or Sigma \cite{Rosenbloom2016}. For the purpose of this paper, ACT-R theory will be the primary lens for discussion on the Common Model. 

\section{Chunks}

Chunks function as the unit of communication between modules in the Common Model, usually conveying propositional information \cite{laird2017standard}. This is exemplified by a chunk such as \textit{name:Fido isa:dog breed:labrador}. Chunks can also be used to request information from long-term, declarative memory \cite{Stewart2007}. For example, \textit{name:? isa:dog breed:labrador} requests the missing value in the first slot of this chunk and will match to any chunk, given the values of the other slots. Chunks can be generated by any module and are communicated between modules by placing them into buffers. The specific structure of chunks holds significance since it is the structure and contents of the chunks that trigger matching productions in the procedural memory. Chunks can be coded as symbolic structures, holographic vectors \citep[that can be unpacked to reveal the chunk structure, see][]{Kelly2020hdm,Kelly2015} or as neural signals \citep[functioning similarly to holographic vectors, see][]{Eliasmith2013}.

\section{Buffers}

 In ACT-R \cite{Anderson1998}, buffers transmit information between the central production system and peripheral modules. Together, all the buffers form the architecture's working memory. Buffering refers to temporarily storing data in a reserved memory space to allow for data processing, transmission, or communication of information between different components or systems. Information enters working memory from different modules, in parallel, and at varying rates. For the central production system to operate within its 50ms cycle, the incoming information needs to be buffered.

\section{Modules and Declarative Memory}

\begin{figure}[t]
\centering
\includegraphics[width=0.9\columnwidth]{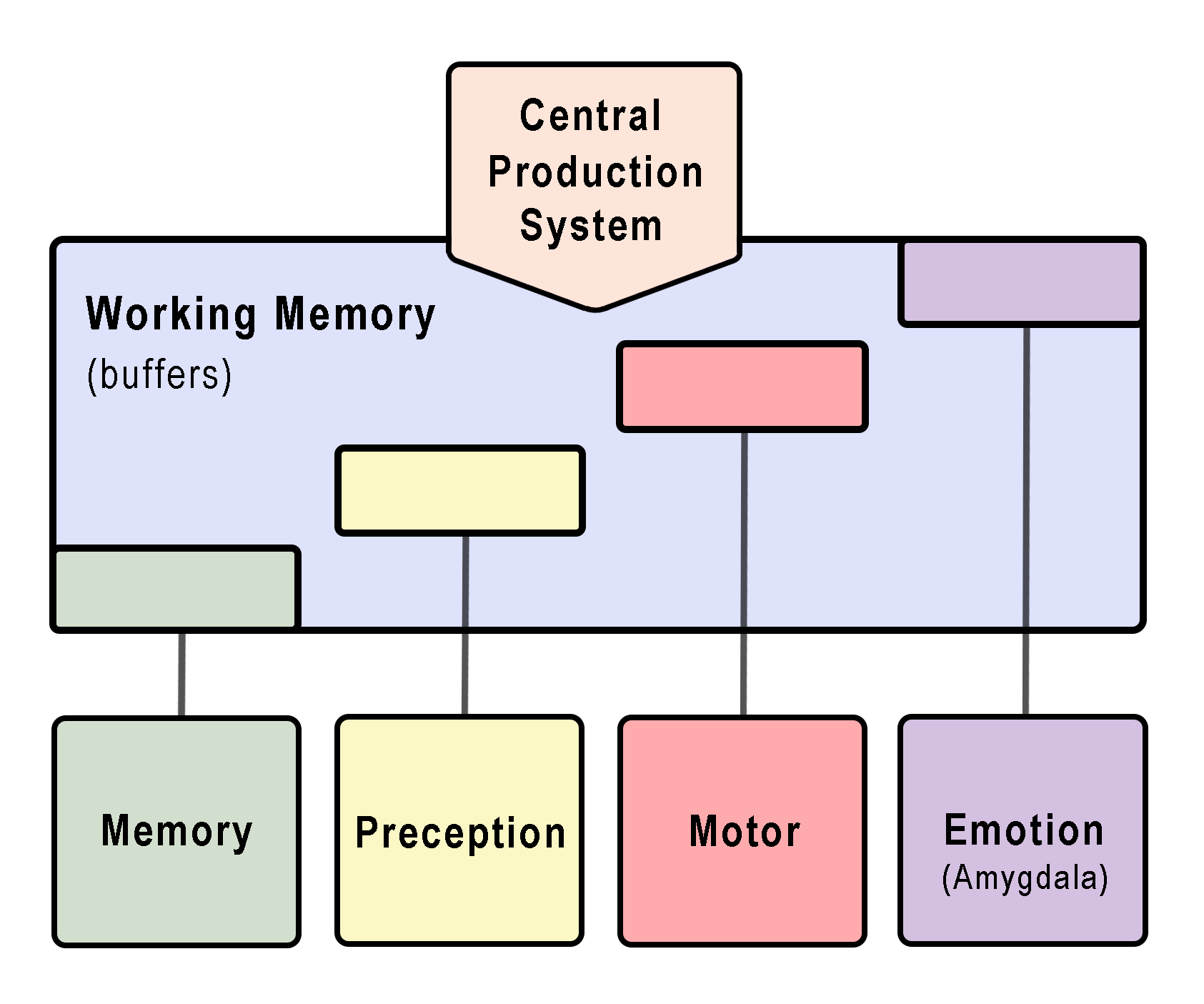} 
\caption{The Common Model Architecture.}
\label{CMC}
\end{figure}

Within the Common Model \cite{laird2017standard}, the peripheral modules represent distinct cognitive and neural functions (see Figure \ref{CMC}). These include the declarative memory module, the perception module, the motor module, and more recently, an emotion module has been proposed. Modules can receive requests from the procedural memory, typically in the form of chunk templates with missing information. When triggered, these requests can lead to visual searches, motor actions, or emotional evaluations. The peripheral module responses are converted into chunks and placed into buffers associated with their respective modules, allowing procedural memory to access the information.

The declarative memory module functions as a warehouse for previously stored chunks, which it can retrieve upon request. The central production system can request information on various topics, and the declarative memory will search for the appropriate chunk and place it in the buffer. As a result of this interplay, cognition can be driven both by productions, which guide actions from predefined rules, and by declarative memory when productions seek guidance from stored information. This combination allows for flexible and dynamic cognitive processes in the Common Model.

\section{Production Rules}

The fundamental structure of a production rule is a conditional-action pairing \cite{Stocco2021cmc}. A rule specifies a condition that, when met, performs a prescribed action. This can also be referred to as an ``if-then'' rule, where the conditional ``if'' side matches to the content of a buffer. Buffers are the working memory of the system, functioning as input and output for the various productions. If the condition it specifies matches the buffer content, then it executes a prescribed action. The actionable ``then" side of the production executes instructions that communicate with peripheral modules or modify the buffer contents. 

In cases where multiple productions match, the production with the highest utility value is selected. Overall, this process constitutes a single cycle of cognition. Although production rules generally refer to pre-existing rules, production rules can also be formed at the time of matching or developed over time. Different Common Model architectures adopt diverse approaches for surpassing fixed production rules \cite{kieras1997epic,laird2012,laird2017standard}.

An agent's current situation is encoded within buffers, and production rules represent the process of selecting between alternative choices \cite{Newell1973}. In some circumstances, choices are rendered by a single production rule, such as in fast-paced video games, where specific signals prompt predetermined, well-learned responses \cite{Greve2020actr}. However, it is more common for multiple production rules to act in sequence to carry out the steps involved in a choice. For example, production rules can choose specific strategies for decision-making (metacognition) or request additional information (memory retrieval). Production rules working in sequence can execute algorithms, logical thinking, causal reasoning, and heuristics.

The primary characteristics of production rules in the CMC lie less within the details of their functioning, and more prominently within three shared properties. First, productions operate in a sequential manner, creating a serial bottleneck where one production fires at a time. This holds true even in cases where claims of parallel production exist, such as in the Epic architecture \cite{kieras1997epic}. Essentially, multiple productions are combined into one larger, single production, while still adhering to a serial bottleneck. 

A second key attribute is timing. The timescale for productions is set at 50ms, which is analogous to the timing of \citeauthor{newell1994unified}'s (\citeyear{newell1994unified}) concept of a deliberate act in the cognitive band, according to his system levels (see Table \ref{levels}). By scaling up logarithmically from general principles of neural timing, \citeauthor{newell1994unified} estimated the timing to be approximately 100ms. The 50ms timing is based on the success of various CMC models of human reaction times, as well as models of the basal ganglia where the procedural memory is believed to be located \cite{senft2016basalganglia,Stewart2010,Stocco2021cmc}.   

The third characteristic of CMC production systems is their correspondence, to a greater or lesser extent, with symbolic processing. In general, productions manipulate chunks of information that align with specific units of knowledge. Although chunks of information can be expressed either as discrete symbols or as continuous-valued vectors and/or neural activity \cite{Kelly2020hdm,Rutledge-Taylor2014,Stewart2010}, the productions that operate on the chunks always impose a binary choice: a production either matches or does not match to the current situation. Hence, ambiguous stimuli produce non-ambiguous responses from the agent. Should the agent see an animal that is either a large dog or a small bear, the agent either approaches the animal or keeps a safe distance, rather than executing a muddled combination of the two.

\section{Shadow Productions}

A shadow production system is, like the central production system, a set of ``if-then'' rules that fire in response to the content of memory and in turn modify the content of memory. However, the shadow productions are located not in the central executive, but within peripheral cognitive modules, such as perception, motor, and emotion. While shadow productions are not an official part of the CMC, they have been used effectively in CMC models \cite{west2017emotion}. They were developed to explain how emotions, particularly from the amygdala, can generate interruptions and influence behaviour at the cognitive level. Shadow productions are an auxiliary system to the central production system. Shadow productions operate in parallel with central productions, yet they do not disrupt the serial bottleneck. Their main role is to monitor signals from lower levels, such as threat detection in the case of the amygdala. Additionally, a shadow production can get information from the working memory buffers to provide context for evaluating a threat. If a shadow production determines the presence of a threat, it places information in a buffer accessible to the central production system for further use.

While shadow productions can potentially match to any buffer, they can only write to one buffer, whereas productions within the central production system can write to multiple buffers. Therefore, the central production system keeps track of the overall task, while shadow productions keep track of various aspects of the task, as represented by the peripheral modules (see Figure \ref{CMC}). Overall, shadow productions manage bottom-up information from the peripheral modules, while the central production system manages tasks, including interruptions, in a top-down way.

\section{Newell’s Levels}

In his system levels scheme, \citet{newell1994unified} noted  that the lowest system levels band for understanding the brain (Table \ref{levels}) is best described by the language and mathematics associated with neurons (e.g., networks, activation, vectors). \citeauthor{newell1994unified} viewed this as different from the system levels within the cognitive band which is best described with the language of computation and choice (e.g., chunks, symbols, algorithms, heuristics). In terms of functionality, one can characterize the function of generative neural networks as that of prediction. However, some theorists go further and claim that all human cognition is a predictive process \citep[i.e., anticipating what comes next and predicting what information may be useful in a certain context;][]{clark2013next,hutchinson2019predictions,parr2022active}. We accept the former, and reject the latter; neural networks perform prediction well, but cognition is more than predictions alone. Namely, cognition requires decision using symbolic reasoning. Descriptions at the cognitive level \citet{newell1994unified} are traditionally characterized by a choice between alternatives using symbolic reasoning, which \citeauthor{newell1994unified} represented in the form of productions. Overall, cognitive level approaches are good for modelling symbolic decision making and neural level approaches are good for modeling non-symbolic prediction. A unified framework can thus avail itself of both the computational advantages of predictive processing implemented at the neural level and symbolic decision-making implemented at the cognitive level.

\begin{table}[t]
\centering
\begin{tabular}{l|l|l|l}
    Scale & Time Units & Level & Band \\
    \hline
    $10^7$ & months &  &  \\
    $10^6$ & weeks &  & \textbf{Social} \\
    $10^5$ & days &  &  \\
    \hline
    $10^4$ & hours & task &  \\
    $10^3$ & 10 min & task & \textbf{Rational} \\
    $10^2$ & minutes & task & \\
    \hline
    $10^1$ & 10 sec & unit task &  \\
    $10^0$ & 1 sec & operations & \textbf{Cognitive} \\
    $10^{-1}$ & 100 ms & deliberate act & \\
    \hline
    $10^{-2}$ & 10 ms & neural circuit &  \\
    $10^{-3}$ & 1 ms & neuron & \textbf{Biological} \\
    $10^{-4}$ & 100 $\mu$s & organelle & \\
    \end{tabular}
\caption{\citet{newell1994unified} system levels and bands for the time scales of human action.}
\label{levels}
\end{table}

In practical terms, we adopt a hybrid approach to describing and coding different parts of the framework as either symbolic or connectionist. Tokenization and detokenization (moving between vectors and symbols) can be handled in a similar manner to systems such as Holographic Declarative Memory \cite{Kelly2020hdm,Kelly2015,KellyReitter2017hdm}, where vectors containing chunked, propositional information can be unpacked into symbolic chunks, and symbolic chunks can be packed into vectors. Note, that the entire system can also be rendered in a neural form (e.g., using NENGO) in which case unpacking would refer to extracting specific vectors from composite vectors. Following \citet{newell1994unified}, we are not focused on the nature of the representation but rather the function. 

As an analogy for how our proposed system operates, consider the use of ChatGPT \cite{chatgpt2022openai}. A human using ChatGPT will provide a prompt and wait for it to produce an output in the form of natural language. ChatGPT first selects the most important words in the prompt, tokenizes them, and strengthens the weight of those tokens at the model’s input. Then it predicts the next tokenized word in its writing. Once the word is selected, it updates the prompt to include what it has written so far and then repeats the cycle. When it has selected its tokenized response, it then detokenizes the output into natural language. Chat GPT does this until it has written something. The human then evaluates the meaning of what was written, accepts it, or asks ChatGPT to redo it, or makes specific edits, or creates better prompts. In this analogy, the human is the symbolic reasoner and ChatGPT is the prediction system. However, note that something is missing. There is a middle space wherein ChatGPT executes specific actions having to do with tokenization, attention, updating, and iterating. These actions are not predictions, rather they are low level instructions (or rules) designed to pull the right response out of the network. Our proposal involves adding this sort of middle space to the Common Model architecture.

\section{Proposed Framework}

\begin{figure}[t]
\centering
\includegraphics[width=0.9\columnwidth]{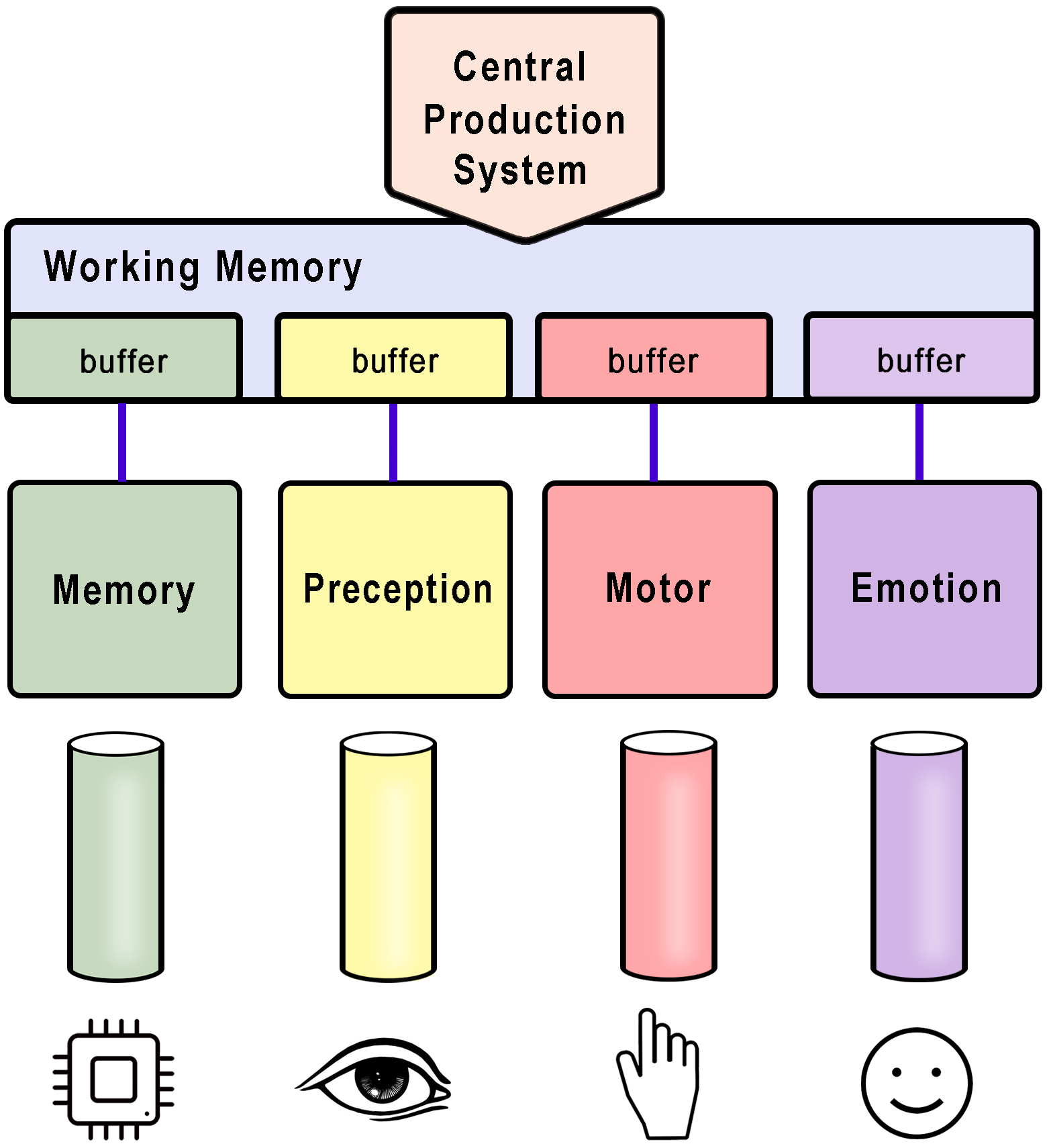} 
\caption{The Pipeline Architecture. Modules in the CMC connect to underlying networks associated with their functionality.}
\label{pipeline}
\end{figure}

\begin{figure}[t]
\centering
\includegraphics[width=0.9\columnwidth]{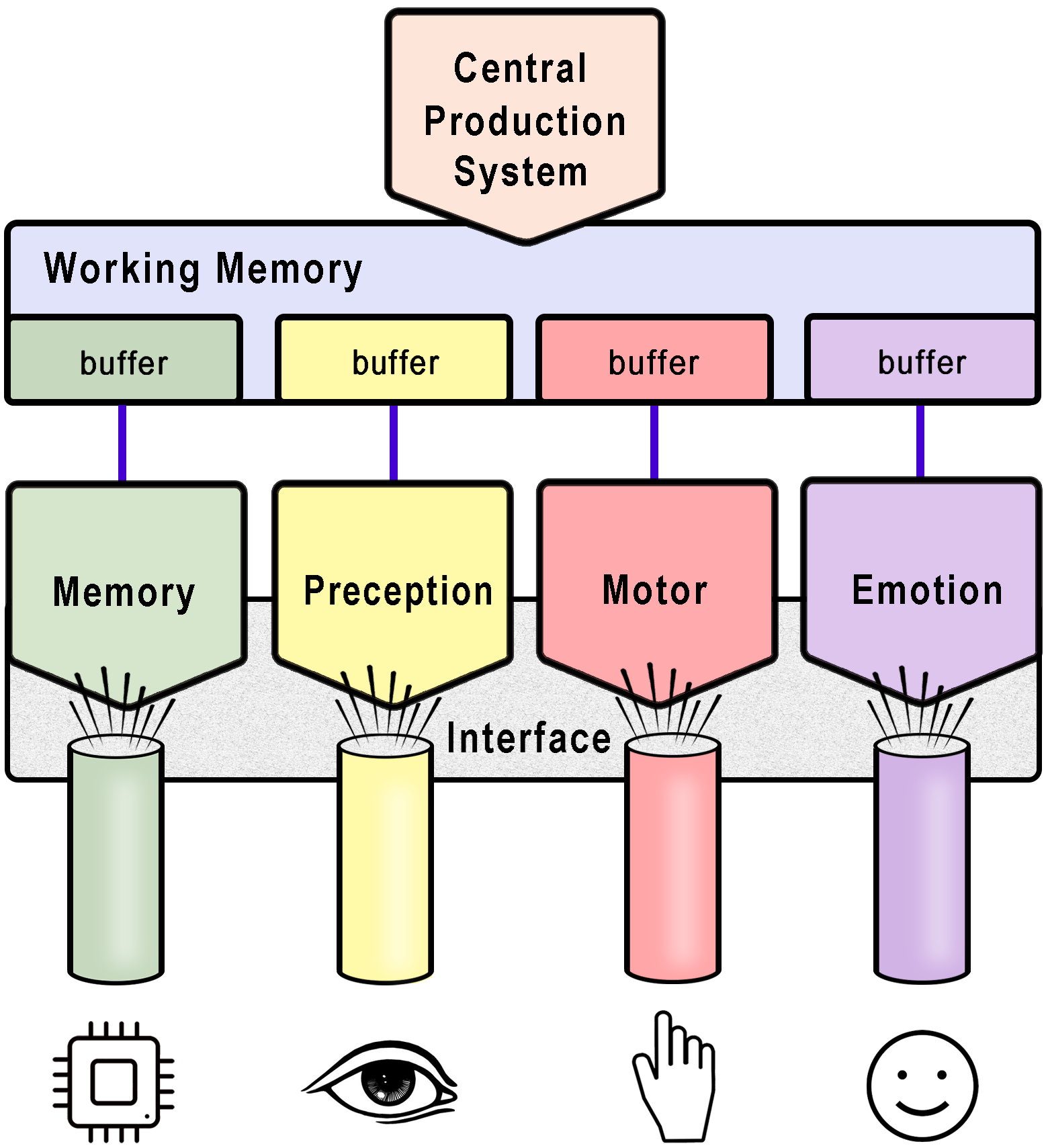} 
\caption{The Middle Memory Architecture. Generative network outputs are tagged with the network that produced them and deposited into the interface (Middle Memory) where they can be accessed by the shadow productions.}
\label{interface}
\end{figure}

The CMC can be connected to generative networks in a straightforward way by having each module of the CMC connect to a corresponding network (see Figure~\ref{pipeline}). The SAL cognitive architecture is a good example of how to integrate ACT-R with a neural network placed inside of a module \cite{Lebiere2008}. However, unless the ``Pipeline" architecture in Figure~\ref{pipeline} is gated or buffered it would put a very heavy load on the serial bottleneck created by the main production system. Our alternative proposal re-conceptualizes the CMC in significant ways (see Figure~\ref{interface}). 

First, we propose an interface, which we call Middle Memory (MM), between the CMC and the various underlying predictive networks. The MM receives vectors from the networks representing the information that the networks predict will be useful (e.g., the next word, in the case of ChatGPT). The MM receives vectors from all modalities (e.g., vectors representing vision, words, emotions, etc.). The MM is similar to Declarative Memory (DM) in ACT-R, and also to Working Memory (WM) in Soar, in that it assigns activation values to the vectors, such that vectors with a higher activation are more likely to be passed to working memory for use by the central production system. Activation also serves as a way to clear out irrelevant information through forgetting \cite{schooler1997rational}. Activation is based on recency, frequency, and spreading activation from WM, which is a much more restricted system of buffers, similar to WM in ACT-R (this is to create human-like constraints). Similar to both Soar WM and ACT-R DM, procedural memory can store propositional chunks forming graph structures in MM (note: we have chosen to mix network predictions and graph structures in MM because we believe there is a potential for synergy, but they could also be stored separately). 

Second, we model all of the CMC modules, except procedural and working memory, as shadow production systems. We propose that these shadow productions fire by matching to information in WM and MM, where WM provides context about the current focus of the system and the MM provides a mixture between predictions of what will be useful (from the underlying networks) and a graph-based understanding of the situation. As noted above, we are not focused on representational issues. Everything in MM could be converted to vectors or everything could be converted to symbolic propositions, or they could be mixed (as the matching specifications for a production can be mixed). We are not saying that the choice of representation makes no difference, just that it has no implications for the level at which we are describing this architecture. 

Information from the different generative networks is tagged with the origin network (e.g., the visual network, the emotional network) when placed into the MM. The different shadow productions match on specific tags, but they are not restricted to only tags from specific origin networks. For example, it may be implemented such that a vision shadow production matches to outputs pertaining to emotion and vision (e.g., a scary animal). This can account for how one modality can affect another in human cognition. The shadow production can also match on elements in WM and elements in the graph generated in MM. Overall, the function of the shadow production systems is to select the best production to fire, given (1) the current focus, (2) the representation of the task, and (3) the predictions of all the networks. Each shadow production system is an expert on a particular modality but also draws on other information to contextualize its outputs. All information placed in the WM buffers is placed there by the shadow productions that fire. No conflicts arise because each shadow production system has its own buffer in WM. Thus, one way to characterize the function of the shadow production is as a way to refine or customize information before allowing it into WM.

\begin{figure}[t]
\centering
\includegraphics[width=0.9\columnwidth]{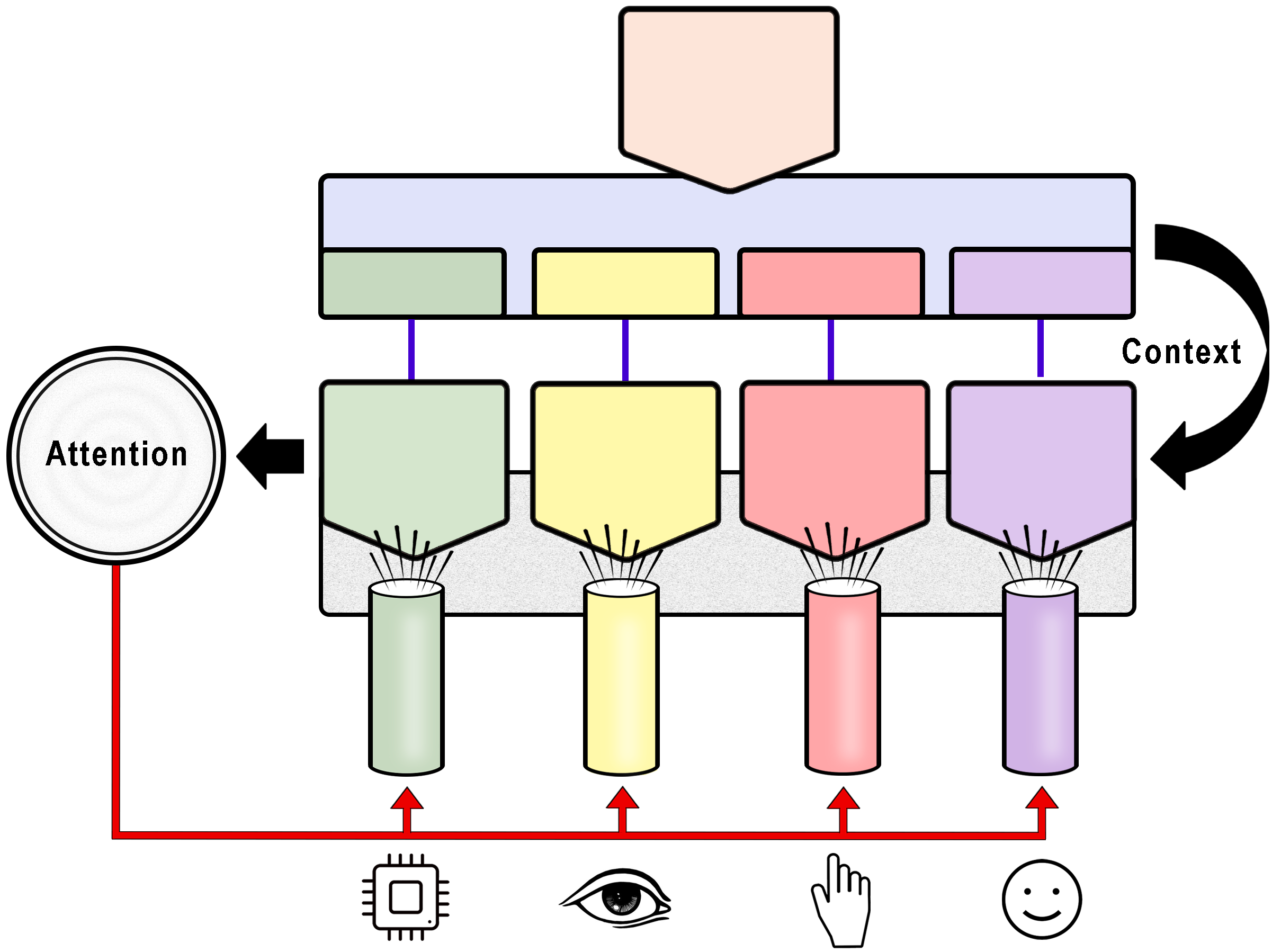} 
\caption{Context and Attention. This depicts how context and attention can feed into the generative networks.}
\label{attention}
\end{figure}

Requests for information from the main production system can be handled by putting queries into the WM buffer associated with the query. Shadow productions would match on the query and whether or not the target information is available in the MM. But, when not responding to requests, the shadow productions would deliver whatever bottom up information they deem useful. This creates an important mechanism for interrupting the main production system in response to unexpected or unusual events  \cite{west2007sgoms}. Notice, though, that requests cannot be made directly to the generative networks. Instead, we propose that the contents of WM are combined with the contents of MM (weighted according to activation) into a single, large vector that is delivered, on every cycle of the main production system, as inputs to all of the generative networks. Figure~\ref{attention} illustrates a general schema for the interaction. The idea is that the generative networks should anticipate what will be required and deliver it.

Production systems that implement Procedural Memory (PM) in the CMC generally use Time Delayed (TD) learning algorithms to adjust the utility values of productions that lead to a reward or punishment. One reason for using production systems for all of the modules is to include them in the TD learning. Our proposal is that when the central production system (i.e., the production system that implements PM) achieves a reward, all of the shadow productions that contributed (by delivering the information to WM that was used to get the reward) would also get a boost to their utility. Another, more difficult issue is, how are the shadow productions created? We propose something similar to Clarion \cite{Sun2017} in which MM contents that have very high activations cause productions to be formed in order to retrieve them. These productions then become permanent if they are rewarded and their utility is boosted (this is similar to the production compilation process in ACT-R). These are very general ideas, and there are numerous ways to do this. However, our point is that bringing all the modules under one learning algorithm could have advantages.

\section{Conclusions}

Humans acquire rules by way of very few repetitions, while neural networks acquire rules through a great many repetitions. Additionally, these learned rules may be susceptible to catastrophic interference, a phenomenon that does not occur with human learning. Humans may forget to apply a rule, yet learning a new rule will not cause previously acquired ones to vanish or change. These observations suggest that human rule acquisition differs greatly from rule acquisition with neural networks. This is not to suggest that the human rule learning system is not neural—it certainly is—however it incorporates special neural structures so these circuits may scale in a way that soft specializations between regions aid in cognition. The basal ganglia \cite{Stewart2010} and the hippocampus are examples of such structures. In the CMC, Procedural Memory is believed to correspond to the basal ganglia \cite{Stewart2010,Stocco2021cmc}. Given the proposed role of Middle Memory, we speculate it may relate to the activity of the thalamus.

This points to a significant division within artificial intelligence research as to whether emergent features within generative networks are sufficient to model the whole of human intelligence, or whether they require structural constraints, such as those provided by the CMC, to produce higher level cognitive processes. We argue that connecting them to the CMC (or similar architectures) is necessary. This can allow for scaling up from statistical, associative learning to more complex reasoning and inference. 

One example where this approach could be useful is causal reasoning. Causal reasoning \cite{pearl2009causality} is the process by which causality is inferred from both the presence and absence of statistical associations, their conditions, and their outcomes. Causal reasoning may be necessary for robust generalization of learning across contexts and is possibly a precondition of human-level intelligence \cite{juliani2022conscious}. Importantly, causal reasoning requires both associative prediction and systematic symbolic thought.

Another example where this approach could be useful is metacognition. The shadow production system can accommodate metacognitive processes \cite{conway2023metathreshold,conway2023metaskill} since summary information of central production activity can be passed to a shadow production system, providing narrative-like information about the model’s own state. Employing this self-referential information, shadow productions can suggest new goals to the central production system. This also means that a metacognitive production system would have limited knowledge of the activity of other shadow production systems, demonstrating a delineation between metacognitive access and more automated aspects of cognition.

More generally, this type of system could be useful for modeling complex forms of expertise, where context is important. In this case, shadow productions can be used to detect network predictions that are not in line with what is expected, and trigger an alarm or cautionary note to the main production system \cite{west2007sgoms}.

In terms of implementation, holographic vectors employ features of both symbolic and connectionist processing and can instantiate the symbols available to a production system as something akin to patterns of neural activity in a declarative memory module. For example, \citeauthor{Kelly2020hdm}'s Holographic Declarative Memory (HDM) uses an algebraic syntax on stored patterns to encode logical connectives on symbols. This is also useful as holographic vectors can insulate networks from the effects of catastrophic interference \cite{Cheung2019superposition,mannering2021catastrophic}. This type of system could definitely be used to build a CMC architecture based solely on vector representations. This would have obvious advantages in connecting to generative networks, which also deal in vectors.

Implementing this proposal into the Common Model of Cognition would provide both a theoretical and practical means of uniting cognitive architectures with generative networks, broadening the scope of human behaviours that can be computationally modelled. By integrating predictive processing with symbolic reasoning in the multimodal context of the CMC, we have proposed an innovative approach to tackle persistently unresolved issues in the field of AI.

\section{Acknowledgments}
We acknowledge the support of the Natural Sciences and Engineering Research Council of Canada (NSERC), [DGECR-2023-00200].

\bibliography{aaai23}

\end{document}